\pdfoutput=1
\documentclass[11pt]{article}

\usepackage{acl}

\usepackage{times}
\usepackage{latexsym}
\usepackage{multirow}
\usepackage{graphicx}
\usepackage{amsmath}
\usepackage{amssymb}
\usepackage{enumitem}
\usepackage[ruled,vlined]{algorithm2e}
\usepackage{subcaption}
\usepackage{makecell}

\usepackage{acl}

\usepackage[T1]{fontenc}
\usepackage[utf8]{inputenc}

\usepackage{microtype}

\title{Global Entity Disambiguation with BERT}

\author{
    Ikuya Yamada$^{\dagger,\ddagger}$\qquad
    Koki Washio$^\mathsection$\thanks{\hspace{2mm}Work done at RIKEN.}\qquad
    Hiroyuki Shindo$^{\mathparagraph,\ddagger}$\qquad
    Yuji Matsumoto$^\ddagger$\\
    $^\dagger$Studio Ousia \quad
    $^\ddagger$RIKEN \quad
    $^\mathsection$Megagon Labs \quad
    $^\mathparagraph$NAIST\\
    \texttt{ikuya@ousia.jp} \quad 
    \texttt{kwashio@megagon.ai} \quad 
    \texttt{shindo@is.naist.jp} \\
    \texttt{yuji.matsumoto@riken.jp} \\ 
}

\begin{document}
\maketitle
\begin{abstract}
  We propose a global entity disambiguation (ED) model based on BERT \cite{devlin2018bert}.
  To capture global contextual information for ED, our model treats not only words but also entities as input tokens, and solves the task by sequentially resolving mentions to their referent entities and using resolved entities as inputs at each step.
  We train the model using a large entity-annotated corpus obtained from Wikipedia.
  We achieve new state-of-the-art results on five standard ED datasets: AIDA-CoNLL, MSNBC, AQUAINT, ACE2004, and WNED-WIKI.
  The source code and model checkpoint are available at \url{https://github.com/studio-ousia/luke}.
\end{abstract}

\section{Introduction}

Entity disambiguation (ED) refers to the task of assigning mentions in a document to corresponding entities in a knowledge base (KB).
This task is challenging because of the ambiguity between mentions (e.g., ``World Cup'') and the entities they refer to (e.g., FIFA World Cup or Rugby World Cup).
ED models typically rely on \textit{local} contextual information based on words that co-occur with the mention and \textit{global} contextual information based on the entity-based coherence of the disambiguation decisions.
A key to improve the performance of ED is to effectively combine both local and global contextual information \cite{ganea-hofmann:2017:EMNLP2017,le-titov-2018-improving}.

\begin{figure}[t]
  \centering
  \includegraphics[width=\columnwidth]{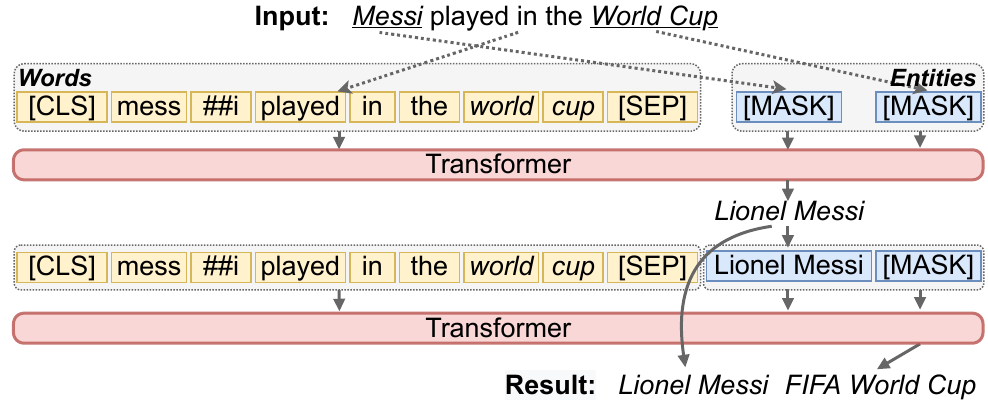}
  \caption{The inference procedure of our model with the input text ``Messi played in the World Cup.''  Given mentions (``Messi'' and ``World Cup''), our model sequentially resolves them to their referent entities, and uses the resolved entities as contexts at each step.}
  \label{fig:inference}
\end{figure}

In this study, we propose a global ED model based on BERT \cite{devlin2018bert}.
Our model treats words and entities in the document as input tokens, and is trained by predicting randomly masked entities in a large entity-annotated corpus obtained from Wikipedia.
This training enables the model to learn how to disambiguate masked entities based on words and non-masked entities.
At the inference time, our model disambiguates mentions sequentially using words and already resolved entities (see Figure \ref{fig:inference}).
This sequential inference effectively accumulates the global contextual information and enhances the coherence of disambiguation decisions \cite{yang2019learning}.

We conducted extensive experiments using six standard ED datasets, i.e., AIDA-CoNLL, MSNBC, AQUAINT, ACE2004, WNED-WIKI, and WNED-CWEB.
As a result, the global contextual information consistently improved the performance.
Furthermore, we achieved new state of the art on all datasets except for WNED-CWEB.
The source code and model checkpoint are available at \url{https://github.com/studio-ousia/luke}.

\begin{figure*}[t]
  \centering
  \includegraphics[width=0.9\textwidth]{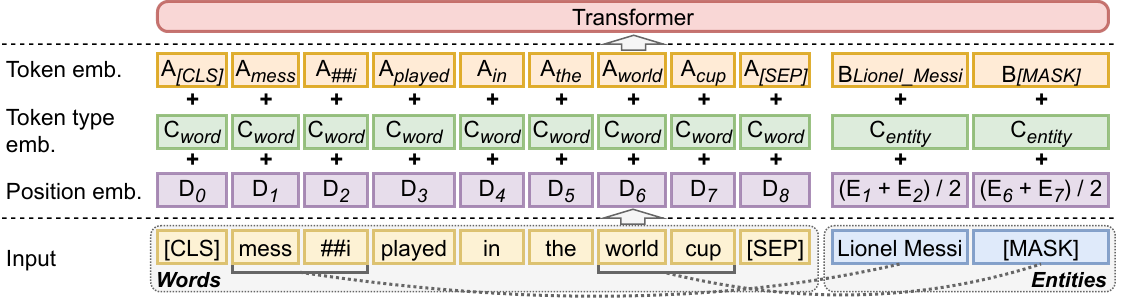}
  \caption{The input representation of our model with the text ``Messi played in the World Cup'' with mentions ``Messi'' and ``World Cup''. The entity corresponding to the mention ``World Cup'' is replaced by the \texttt{[MASK]} token.}
  \label{fig:input-representations}
\end{figure*}

\section{Related Work}
\label{sec:related-work}

\paragraph{Transformer-based ED.} Several recent studies have proposed ED models based on Transformer \cite{NIPS2017_7181} trained with a large entity-annotated corpus obtained from Wikipedia \cite{Broscheit2019InvestigatingLinking,Ling2020LearningText,fevry2020empirical,cao2021autoregressive,barba-et-al-2022}.
\newcite{Broscheit2019InvestigatingLinking} trained an ED model based on BERT by classifying each word in the document to the corresponding entity.
Similarly, \newcite{fevry2020empirical} addressed ED using BERT by classifying mention spans to the corresponding entities.
\newcite{Ling2020LearningText} trained BERT by predicting entities using the document-level representation.
\newcite{cao2021autoregressive} addressed ED by training BART \cite{Lewis2020} to generate referent entity titles of target mentions in an autoregressive manner.
\newcite{barba-et-al-2022} formulated ED as a text extraction problem; they fed the document and candidate entity titles to BART and Longformer \cite{beltagy2020longformer} and disambiguated a mention in the document by extracting the referent entity title of the mention.
However, unlike our model, these models addressed the task based only on local contextual information.

\paragraph{Treating entities as inputs of Transformer.}
Recent studies \cite{Zhang2019,yamada-etal-2020-luke,sun-etal-2020-colake} have proposed Transformer-based models that treat entities as input tokens to enrich their expressiveness using additional information contained in the entity embeddings.
However, these models were designed to solve general NLP tasks and not tested on ED.
We treat entities as input tokens to capture the global context that is shown to be highly effective for ED.

\paragraph{ED as sequential decision task.} Past studies \cite{yang2019learning,Fang:2019:JEL:3308558.3313517} have solved ED by casting it as a sequential decision task to capture global contextual information.
We adopt a similar method with an enhanced Transformer architecture, a training task, and an inference method to implement the global ED model based on BERT.

\section{Model}
\label{sec:embedding-model}

Given a document with $N$ mentions, each of which has $K$ entity candidates, our model solves ED by selecting a correct referent entity from the entity candidates for each mention.

\subsection{Model Architecture}

Our model is based on BERT and takes words and entities (Wikipedia entities or the \texttt{[MASK]} entity).
The input representation of a word or an entity is constructed by summing the token, token type, and position embeddings (see Figure \ref{fig:input-representations}):

\vspace{1mm}
\noindent {\bf Token embedding} is the embedding of the corresponding token.
The matrices of the word and entity token embeddings are represented as $\mathbf{A} \in \mathbb{R}^{V_{w}\times H}$ and $\mathbf{B} \in \mathbb{R}^{V_{e}\times H}$, respectively, where $H$ is the size of the hidden states of BERT, and $V_{w}$ and $V_{e}$ are the number of items in the word vocabulary and that of the entity vocabulary, respectively.

\vspace{1mm}
\noindent {\bf Token type embedding} represents the type of token, namely word ($\mathbf{C}_{word}$) or entity ($\mathbf{C}_{entity}$).

\vspace{1mm}
\noindent {\bf Position embedding} represents the position of the token in a word sequence.
A word and an entity appearing at the $i$-th position in the sequence are represented as $\mathbf{D}_i$ and $\mathbf{E}_i$, respectively.
If an entity mention contains multiple words, its position embedding is computed by averaging the embeddings of the corresponding positions (see Figure \ref{fig:input-representations}).

Following \newcite{devlin2018bert}, we tokenize the document text using the BERT's wordpiece tokenizer, and insert \texttt{[CLS]} and \texttt{[SEP]} tokens as the first and last words, respectively.

\subsection{Training Task}
\label{subsec:mep}

Similar to the masked language model (MLM) objective adopted in BERT, our model is trained by predicting randomly masked entities.
Specifically, we randomly replace some percentage of the entities with special \texttt{[MASK]} entity tokens and then trains the model to predict masked entities.

We adopt a model equivalent to the one used to predict words in MLM.
Formally, we predict the original entity corresponding to a masked entity by applying softmax over all entities:
\begin{align}
  \mathbf{\hat{y}} & = \text{softmax}(\mathbf{B}\mathbf{m}_e + \mathbf{b}_{o}) \label{eq:entity-prediction}              \\
  \mathbf{m}_e     & = \text{layernorm}\big(\text{gelu}(\mathbf{W}_f\mathbf{h}_e + \mathbf{b}_f)\big) \label{eq:feature}
\end{align}
where $\mathbf{h}_e \in \mathbb{R}^H$ is the output embedding corresponding to the masked entity, $\mathbf{W}_f \in \mathbb{R}^{H\times H}$ is a matrix, $\mathbf{b}_{o} \in \mathbb{R}^{V_e}$ and $\mathbf{b}_f \in \mathbb{R}^H$ are bias vectors, $\text{gelu}(\cdot)$ is the gelu activation function \cite{hendrycks2016gaussian}, and $\text{layernorm}(\cdot)$ is the layer normalization function \cite{lei2016layer}.

\subsection{ED Model}

\paragraph{Local ED Model.}

Our local ED model takes words and $N$ \texttt{[MASK]} tokens corresponding to the mentions in the document.
The model then computes the embedding $\mathbf{m}'_e \in \mathbb{R}^{H}$ for each \texttt{[MASK]} token using Eq.\eqref{eq:feature} and predicts the entity using softmax over the $K$ entity candidates:
\begin{equation}
  \label{eq:ed-prediction}
  \mathbf{\hat{y}}_{\scriptscriptstyle{ED}} = \text{softmax}(\mathbf{B}^*\mathbf{m}'_e + \mathbf{b}^*_o),
\end{equation}
where $\mathbf{B}^* \in \mathbb{R}^{K\times H}$ and $\mathbf{b}_o^* \in \mathbb{R}^K$ consist of the entity token embeddings and the bias corresponding to the entity candidates, respectively.
Note that $\mathbf{B}^*$ and $\mathbf{b}_o^*$ are the subsets of $\mathbf{B}$ and $\mathbf{b}_o$, respectively.

\paragraph{Global ED Model.}
\label{subsec:global-ed}

\SetAlFnt{\small}
\SetAlCapFnt{\small}
\SetAlCapNameFnt{\small}
\begin{algorithm}[t]
  \SetKwFor{RepTimes}{repeat}{times}{end}
  \SetAlgoLined
  \SetArgSty{}
  \SetKwInput{KwInit}{Initialize}
  \DontPrintSemicolon
  \KwIn{Words and mentions $m_1, \ldots m_N$.}
  \KwInit{$e_i\gets\text{\texttt{[MASK]}}, i = 1 \ldots N$\;}
  \RepTimes{$N$}{
    For all \texttt{[MASK]}s, obtain predictions using Eq.\eqref{eq:ed-prediction} with words and entities $e_1$, ..., $e_N$ as inputs\;
    Select a mention $m_j$ and its prediction $\hat{e}_j$ with the highest probability\;
    $e_j\gets \hat{e}_j$
  }
  \Return{\{$e_1, \ldots, e_N\}$}
  \caption{Algorithm of our global ED model.}
  \label{alg:global-ed}
\end{algorithm}

Our global ED model resolves mentions sequentially for $N$ steps (see Algorithm \ref{alg:global-ed}).
First, the model initializes the entity of each mention using the \texttt{[MASK]} token.
Then, for each step, it predicts an entity for each \texttt{[MASK]} token, selects the prediction with the highest probability produced by the softmax function in Eq.\eqref{eq:ed-prediction}, and resolves the corresponding mention by assigning the predicted entity to it.
This model is denoted as \textbf{confidence-order}.
We also test a model that selects mentions according to their order of appearance in the document and denote it by \textbf{natural-order}.

\begin{table}[t]
  \centering
  \scalebox{0.82}{
    \begin{tabular}{l|c|c}
      \hline
      Name                                                    & \makecell{Accuracy                 \\\small{(KB+YAGO)}}               & \makecell{Accuracy\\\small{(PPRforNED)}} \\
      \hline
      \textbf{Baselines:}                                     &                    &               \\
      \hspace{2mm}\newcite{Yamada2016}                        & 91.5               & 93.1          \\
      \hspace{2mm}\newcite{ganea-hofmann:2017:EMNLP2017}      & 92.2               & -             \\
      \hspace{2mm}\newcite{yang-etal-2018-collective}         & 93.0               & 95.9          \\
      \hspace{2mm}\newcite{le-titov-2018-improving}           & 93.1               & -             \\
      \hspace{2mm}\newcite{Fang:2019:JEL:3308558.3313517}     & 94.3               & -             \\
      \hspace{2mm}\newcite{yang2019learning}                  & 94.6                               \\
      \hspace{2mm}\newcite{Broscheit2019InvestigatingLinking} & 87.9               & -             \\
      \hspace{2mm}\newcite{Ling2020LearningText}              & -                  & 94.9          \\
      \hspace{2mm}\newcite{fevry2020empirical}                & 92.5               & 96.7          \\
      \hspace{2mm}\newcite{cao2021autoregressive}             & 93.3               & -             \\
      \hspace{2mm}\newcite{barba-et-al-2022}             & 92.6               & -             \\
      \hline
      \textbf{Our model w/o fine-tuning:}                     &                    &               \\
      \hspace{2mm}confidence-order                            & 92.4               & 94.6          \\
      \hspace{2mm}natural-order                               & 91.7               & 94.0          \\
      \hspace{2mm}local                                       & 90.8               & 94.0          \\
      \hline
      \textbf{Our model w/ fine-tuning:}                      &                    &               \\
      \hspace{2mm}confidence-order                            & \textbf{95.0}      & \textbf{97.1} \\
      \hspace{2mm}natural-order                               & 94.8               & 97.0          \\
      \hspace{2mm}local                                       & 94.5               & 96.8          \\
      \hline
    \end{tabular}
  }
  \caption{In-KB accuracy on the CoNLL dataset.}
  \label{tb:conll-results}

\end{table}
\subsection{Modeling Details}
\label{subsec:training}

Our model is based on BERT$_{\text{LARGE}}$ \cite{devlin2018bert}.
The parameters shared with BERT are initialized using BERT, and the other parameters are initialized randomly.
We treat the hyperlinks in Wikipedia as entity annotations and randomly mask 30\% of all entities.
We train the model by maximizing the log likelihood of entity predictions.
Further details are described in Appendix \ref{sec:pretraining-details}.

\section{Experiments}
\label{subsec:experimental-setup}

\begin{table*}[tb]
  \centering
  \scalebox{0.82}{
    \begin{tabular}{l|c|c|c|c|c|c}
      \hline
      Name                                                & MSNBC         & AQUAINT       & ACE2004       & CWEB          & WIKI          & Average \\
      \hline
      \textbf{Baselines:}                                 &               &               &               &               &               &         \\
      \hspace{2mm}\newcite{ganea-hofmann:2017:EMNLP2017}  & 93.7          & 88.5          & 88.5          & 77.9          & 77.5 & 85.2                   \\
      \hspace{2mm}\newcite{yang-etal-2018-collective}     & 92.6          & 89.9          & 88.5          & \textbf{81.8} & 79.2 & 86.4                   \\
      \hspace{2mm}\newcite{le-titov-2018-improving}       & 93.9          & 88.3          & 89.9          & 77.5          & 78.0  & 85.5                  \\

      \hspace{2mm}\newcite{Fang:2019:JEL:3308558.3313517} & 92.8          & 87.5          & 91.2          & 78.5          & 82.8  & 86.6                  \\
      \hspace{2mm}\newcite{yang2019learning}              & 93.8          & 88.3          & 90.1          & 75.6          & 78.8  & 85.3                  \\
      \hspace{2mm}\newcite{cao2021autoregressive}         & 94.3          & 89.9          & 90.1          & 77.3          & 87.4  & 87.8                  \\
      \hspace{2mm}\newcite{barba-et-al-2022}         & 94.7          & 91.6          & 91.8          & 77.7          & 88.8  & 88.9                  \\
      \hline
      \textbf{Our model w/o fine-tuning:}                 &               &               &               &               &               &         \\
      \hspace{2mm}confidence-order                        & \textbf{96.3} & \textbf{93.5} & \textbf{91.9} & 78.9          & 89.1 & \textbf{89.9}                   \\
      \hspace{2mm}natural-order                           & 96.1          & 92.9          & \textbf{91.9} & 78.4          & \textbf{89.2} &89.7          \\
      \hspace{2mm}local                                   & 96.1          & 91.9          & \textbf{91.9} & 78.4          & 88.8             &89.4       \\
      \hline
      \textbf{Our model w/ fine-tuning:}                  &               &               &               &               &               &         \\
      \hspace{2mm}confidence-order                        & 94.1          & 91.5          & 90.7          & 78.3          & 87.6  & 88.4                  \\
      \hspace{2mm}natural-order                           & 94.1          & 90.9          & 90.7          & 78.3          & 87.4 &88.3                   \\
      \hspace{2mm}local                                   & 94.1          & 90.8          & 90.7          & 78.2          & 87.2 &88.2                   \\
      \hline
    \end{tabular}
  }
  \caption{Micro F1 score on the MSNBC, AQUAINT, ACE2004, CWEB, and WIKI datasets.}
  \label{tb:results}
\end{table*}

\begin{table}[tb]
  \centering
  \setlength{\tabcolsep}{2pt}
  \scalebox{0.77}{
    \begin{tabular}{c|c|c|c|c}
      \hline
      \#annotations & confidence-order & natural-order & local & \makecell{G\&H2017} \\
      \hline
      0              & 1.0              & 1.0           & 1.0   & 0.8                 \\
      1--10          & 95.55            & 95.55         & 95.55 & 91.93               \\
      11--50         & 96.98            & 96.70         & 96.43 & 92.44               \\
      $\geq$51       & 96.64            & 96.38         & 95.80 & 94.21               \\
      \hline
    \end{tabular}
  }
  \caption{Accuracy on the CoNLL dataset split by the frequency of entity annotations. Our models were fine-tuned using the CoNLL dataset. \textbf{G\&H2017}: The results of \newcite{ganea-hofmann:2017:EMNLP2017}.}
  \label{tb:analysis}
\end{table}

Our experimental setup follows \newcite{le-titov-2018-improving}.
In particular, we test the proposed ED models using six standard datasets: AIDA-CoNLL (CoNLL) \cite{Hoffart2011}, MSNBC, AQUAINT, ACE2004, WNED-CWEB (CWEB), and WNED-WIKI (WIKI) \cite{guo2018robust}.
We consider only the mentions that refer to valid entities in Wikipedia.
For all datasets, we use the \textit{KB+YAGO} entity candidates and their associated $\hat{p}(e|m)$ \cite{ganea-hofmann:2017:EMNLP2017}, and use the top 30 candidates based on $\hat{p}(e|m)$.
For the CoNLL dataset, we also test the performance using \textit{PPRforNED} entity candidates \cite{pershina-he-grishman:2015:NAACL-HLT}.
We report the in-KB accuracy for the CoNLL dataset and the micro F1 score (averaged per mention) for the other datasets.
Further details of the datasets are provided in Appendix \ref{sec:dataset-details}.

Furthermore, we optionally fine-tune the model by maximizing the log likelihood of the ED predictions ($\mathbf{\hat{y}}_{\scriptscriptstyle{ED}}$) using the training set of the CoNLL dataset with the KB+YAGO candidates.
We mask 90\% of the mentions and fix the entity token embeddings ($\mathbf{B}$ and $\mathbf{B}^*$) and the bias ($\mathbf{b}_o$ and $\mathbf{b}_o^*$).
The model is trained for two epochs using AdamW.
Additional details are provided in Appendix \ref{sec:fine-tuning-details}.

\subsection{Results}
Table \ref{tb:conll-results} and Table \ref{tb:results} present our experimental results.
We achieve new state of the art on all datasets except the CWEB dataset by outperforming strong Transformer-based ED models, i.e, \newcite{Broscheit2019InvestigatingLinking}, \newcite{Ling2020LearningText}, \newcite{fevry2020empirical}, \newcite{cao2021autoregressive}, and \newcite{barba-et-al-2022}.\footnote{All models listed in Table \ref{tb:results} use Wikipedia as training data which partly overlap with the WIKI dataset.}
Furthermore, on the CoNLL dataset, our confidence-order model trained only on our Wikipedia-based corpus outperforms \newcite{Yamada2016} and \newcite{ganea-hofmann:2017:EMNLP2017} trained on its in-domain training set.

Our global models consistently perform better than the local model, demonstrating the effectiveness of using global contextual information even if local contextual information is captured using expressive BERT model.
Moreover, the confidence-order model performs better than the natural-order model on most datasets.
An analysis investigating why the confidence-order model outperforms the natural-order model is provided in the next section.

The fine-tuning on the CoNLL dataset significantly improves the performance on this dataset (Table \ref{tb:conll-results}).
However, it generally degrades the performance on the other datasets (Table \ref{tb:results}).
This suggests that Wikipedia entity annotations are more suitable than the CoNLL dataset to train general-purpose ED models.

Additionally, our models perform worse than \newcite{yang-etal-2018-collective} on the CWEB dataset.
This is because this dataset is significantly longer on average than other datasets, i.e., approximately 1,700 words per document on average, which is more than three times longer than the 512-word limit that can be handled by BERT-based models including ours.
\newcite{yang-etal-2018-collective} achieved excellent performance on this dataset because their model uses various hand-engineered features capturing document-level contextual information.

\subsection{Analysis}
To investigate how global contextual information helps our model to improve performance, we manually analyze the difference between the predictions of the local, natural-order, and confidence-order models.
We use the fine-tuned model using the CoNLL dataset with the YAGO+KB candidates.
Although all models perform well on most mentions, the local model often fails to resolve mentions of common names referring to specific entities (e.g., ``New York'' referring to New York Knicks).
Global models are generally better to resolve such difficult cases because of the presence of strong global contextual information (e.g., mentions referring to basketball teams).

Furthermore, we find that the confidence-order model works especially well for mentions that require a highly detailed context to resolve.
For example, a mention of ``Matthew Burke'' can refer to two different former Australian rugby players.
Although the local and natural-order models incorrectly resolve this mention to the player who has the larger number of occurrences in our Wikipedia-based corpus, the confidence-order model successfully resolves this by disambiguating its contextual mentions, including his teammates, in advance.
We provide detailed inference sequence of the corresponding document in Appendix \ref{sec:inference-example}.

\subsection{Performance for Rare Entities}
We examine whether our model learns effective embeddings for rare entities using the CoNLL dataset.
Following \newcite{ganea-hofmann:2017:EMNLP2017}, we use the mentions of which entity candidates contain their gold entities and measure the performance by dividing the mentions based on the frequency of their entities in the Wikipedia annotations used to train the embeddings.

As presented in Table \ref{tb:analysis}, our models achieve enhanced performance for rare entities.
Furthermore, the global models consistently outperform the local model both for rare and frequent entities.

\section{Conclusion and Future Work}
We propose a new global ED model based on BERT.
Our extensive experiments on a wide range of ED datasets demonstrate its effectiveness.

One limitation of our model is that, similar to existing ED models, our model cannot handle entities that are not included in the vocabulary.
In our future work, we will investigate the method to compute the embeddings of such entities using a post-hoc training with an extended vocabulary \cite{tai-etal-2020-exbert}.    

\bibliography{references}
\bibliographystyle{acl_natbib}

\clearpage
\appendix

\section*{Appendix for ``Global Entity Disambiguation with BERT''}

\section{Details of Proposed Model}
\label{sec:pretraining-details}
As the input corpus for training our model, we use the December 2018 version of Wikipedia, comprising approximately 3.5 billion words and 11 million entity annotations.
We generate input sequences by splitting the content of each page into sequences comprising $\leq 512$ words and their entity annotations (i.e., hyperlinks).
The input text is tokenized using BERT's tokenizer with its vocabulary consisting of $V_w = 30,000$ words.
Similar to \newcite{ganea-hofmann:2017:EMNLP2017}, we create an entity vocabulary consisting of $V_e = 128,040$ entities, which are contained in the entity candidates in the datasets used in our experiments.

Our model consists of approximately 440 million parameters.
To reduce the training time, the parameters that are shared with BERT are initialized using BERT.
The other parameters are initialized randomly.
The model is trained via iterations over Wikipedia pages in a random order for seven epochs.
To stabilize the training, we update only those parameters that are randomly initialized (i.e., fixed the parameters initialized using BERT) at the first epoch, and update all parameters in the remaining six epochs.
We implement the model using PyTorch \cite{NEURIPS2019_bdbca288} and Hugging Face Transformers \cite{wolf-etal-2020-transformers}, and the training takes approximately ten days using eight Tesla V100 GPUs.
We optimize the model using AdamW.
The hyper-parameters used in the training are detailed in Table  \ref{tb:pretraining-config}.

\begin{table}[t]
  \centering
  \scalebox{0.85}{
  \begin{tabular}{l|c}
    \hline
    Name                            & Value  \\
    \hline
    number of hidden layers         & 24     \\
    hidden size                     & 1024   \\
    attention heads                 & 16     \\
    attention head size             & 64     \\
    activation function             & gelu   \\
    maximum word length             & 512    \\
    batch size                      & 2048   \\
    learning rate (1st epoch)       & 5e-4   \\
    learning rate decay (1st epoch) & none   \\
    warmup steps (1st epoch)        & 1000   \\
    learning rate                   & 5e-5   \\
    learning rate decay             & linear \\
    warmup steps                    & 1000   \\
    dropout                         & 0.1    \\
    weight decay                    & 0.01   \\
    gradient clipping               & 1.0    \\
    adam $\beta_1$                  & 0.9    \\
    adam $\beta_2$                  & 0.999  \\
    adam $\epsilon$                 & 1e-6   \\
    \hline
  \end{tabular}
  }
  \caption{Hyper-parameters used for training on Wikipedia entity annotations.}
  \label{tb:pretraining-config}
  
\end{table}

\section{Details of Fine-tuning on CoNLL Dataset}
\label{sec:fine-tuning-details}
The hyper-parameters used in the fine-tuning on the CoNLL dataset are detailed in Table \ref{tb:fine-tuning-config}.
We select these hyper-parameters from the search space described in \newcite{devlin2018bert} based on the accuracy on the development set of the CoNLL dataset.
A document is split if it is longer than 512 words, which is the maximum word length of the BERT model.

\begin{table}[t]
  \centering
  \scalebox{0.85}{
  \begin{tabular}{l|c}
    \hline
    Name                & Value  \\
    \hline
    maximum word length & 512    \\
    number of epochs    & 2      \\
    batch size          & 16     \\
    learning rate       & 2e-5   \\
    learning rate decay & linear \\
    warmup proportion   & 0.1    \\
    dropout             & 0.1    \\
    weight decay        & 0.01   \\
    gradient clipping   & 1.0    \\
    adam $\beta_1$      & 0.9    \\
    adam $\beta_2$      & 0.999  \\
    adam $\epsilon$     & 1e-6   \\
    \hline
  \end{tabular}
  }
  \caption{Hyper-parameters during fine-tuning on the CoNLL dataset.}
  \label{tb:fine-tuning-config}
\end{table}

\section{Details of ED Datasets}
\label{sec:dataset-details}

The statistics of the ED datasets used in our experiments are provided in Table \ref{tb:dataset-stats}.

\begin{table}
  \centering
  \centering
  \scalebox{0.85}{
  \begin{tabular}{l|cc}
    \hline
    Name                & \#mentions & \#documents \\
    \hline
    CoNLL (training) & 18,448 & 946 \\
    CoNLL (development) & 4,791  & 216 \\
    CoNLL (test) & 4,485  & 231 \\
    MSNBC & 656  & 20 \\
    AQUAINT & 727  & 50 \\
    ACE2004 & 257  & 36 \\
    CWEB & 11,154  & 320 \\
    WIKI & 6,821  & 320 \\
    \hline
  \end{tabular}
  }
  \caption{Statistics of ED datasets.}
  \label{tb:dataset-stats}
\end{table}

\section{Example of Inference by Confidence-order Model}
\label{sec:inference-example}

\begin{figure*}[tbhp]
  \centering
  \fbox{
    \includegraphics[width=0.975\textwidth]{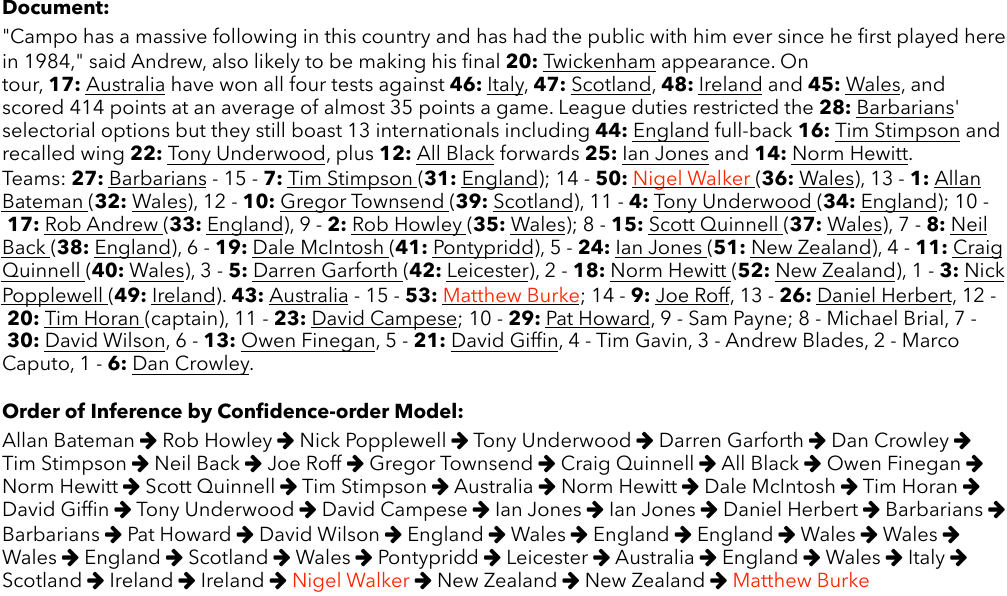}
  }
  \caption{An illustrative example showing the inference performed by our fine-tuned confidence-order model on a document in the CoNLL dataset. Mentions are shown as underlined. Numbers in boldface represent the selection order of the confidence-order model.}
  \label{fig:inference-example}
\end{figure*}

Figure \ref{fig:inference-example} shows an example of the inference performed by our confidence-order model fine-tuned on the CoNLL dataset.
The document is obtained from the test set of the CoNLL dataset.
As shown in the figure, the model starts with unambiguous player names to recognize the topic of the document, and subsequently resolves the mentions that are challenging to resolve.

Notably, the model correctly resolves the mention ``Nigel Walker'' to the corresponding former rugby player instead of a football player, and the mention ``Matthew Burke'' to the correct former Australian rugby player born in 1973 instead of the former Australian rugby player born in 1964. This is accomplished by resolving other contextual mentions, including their colleague players, in advance.
These two mentions are denoted in red in the figure.
Note that our local model fails to resolve both mentions, and our natural-order model fails to resolve ``Matthew Burke.''

\end{document}